\documentclass[letterpaper, 10 pt, conference]{ieeeconf}  

\IEEEoverridecommandlockouts                              

\usepackage{graphics} 
\usepackage{epsfig} 
\usepackage{amsmath} 
\usepackage{amssymb}  
\usepackage{hyperref}
\usepackage{booktabs}
\usepackage{cite}
\usepackage[dvipsnames]{xcolor}

\newcommand{\methodnamenosp}{AutoSelect}
\newcommand{\methodname}{\methodnamenosp~}

\title{\LARGE \bf
\methodnamenosp: Automatic and Dynamic Detection Selection \\for 3D Multi-Object Tracking
\vspace{-0.2cm}
}

\author{Xinshuo Weng and Kris Kitani  
\thanks{Xinshuo Weng and Kris Kitani are with Robotics Institute, Carnegie Mellon University, USA. {\tt\small \{xinshuow, kkitani\}@cs.cmu.edu}.}
\vspace{-1cm}
}

\begin{document}

\maketitle
\thispagestyle{empty}
\pagestyle{empty}

\begin{abstract}

3D multi-object tracking is an important component in robotic perception systems such as self-driving vehicles. Recent work follows a tracking-by-detection pipeline, which aims to match past tracklets with detections in the current frame. To avoid matching with false positive detections, prior work filters out detections with low confidence scores via a threshold. However, finding a proper threshold is non-trivial, which requires extensive manual search via ablation study. Also, this threshold is sensitive to many factors such as target object category so we need to re-search the threshold if these factors change. To ease this process, we propose to automatically select high-quality detections and remove the efforts needed for manual threshold search. Also, prior work often uses a single threshold per data sequence, which is sub-optimal in particular frames or for certain objects. Instead, we dynamically search threshold per frame or per object to further boost performance. Through experiments on KITTI and nuScenes, our method can filter out $45.7\%$ false positives while maintaining the recall, achieving new S.O.T.A. performance and removing the need for manually threshold tuning. 

\end{abstract}


\section{Introduction}

Multi-object tracking (MOT) is one of the essential techniques in modern perception systems and has been widely used in robotic applications such as autonomous driving \cite{Wang2018, Badue2019, Weng2020_SPF2} and assistive robots \cite{Sun2020_VIPL, Manglik2019, Kayukawa2019}. Compared to 2D MOT \cite{Wang2020_GNNDetTrk, Sharma2018, Lee2016, Weng2018_R2N, Ishioka2020_WDTA, Osep2018} which detects/tracks objects in video, 3D MOT \cite{Weng2020_AB3DMOT, Weng2020_GNN3DMOT, Weng2020_GNNTrkForecast} tracks objects in 3D space represented as 3D bounding boxes. To approach 3D MOT, recent work often follows a tracking-by-detection pipeline, where the goal is to match past tracklets with 3D detections in the current frame. As the detections come from real-world 3D object detectors \cite{Weng2019_Mono3DPLiDAR, Shi2019, Yi2020}, there will be inevitably many false positives, which can be wrongly matched to past tracklets or used to create false positive tracklets, leading to sub-optimal performance. Therefore, to achieve strong 3D MOT performance, a detection selection technique is needed to filter out false positives while maintain as many true positives as possible. 

\begin{figure}[t]
\begin{center}
\includegraphics[trim=0.3cm 1.8cm 0.7cm 0cm, clip=true, width=\linewidth]{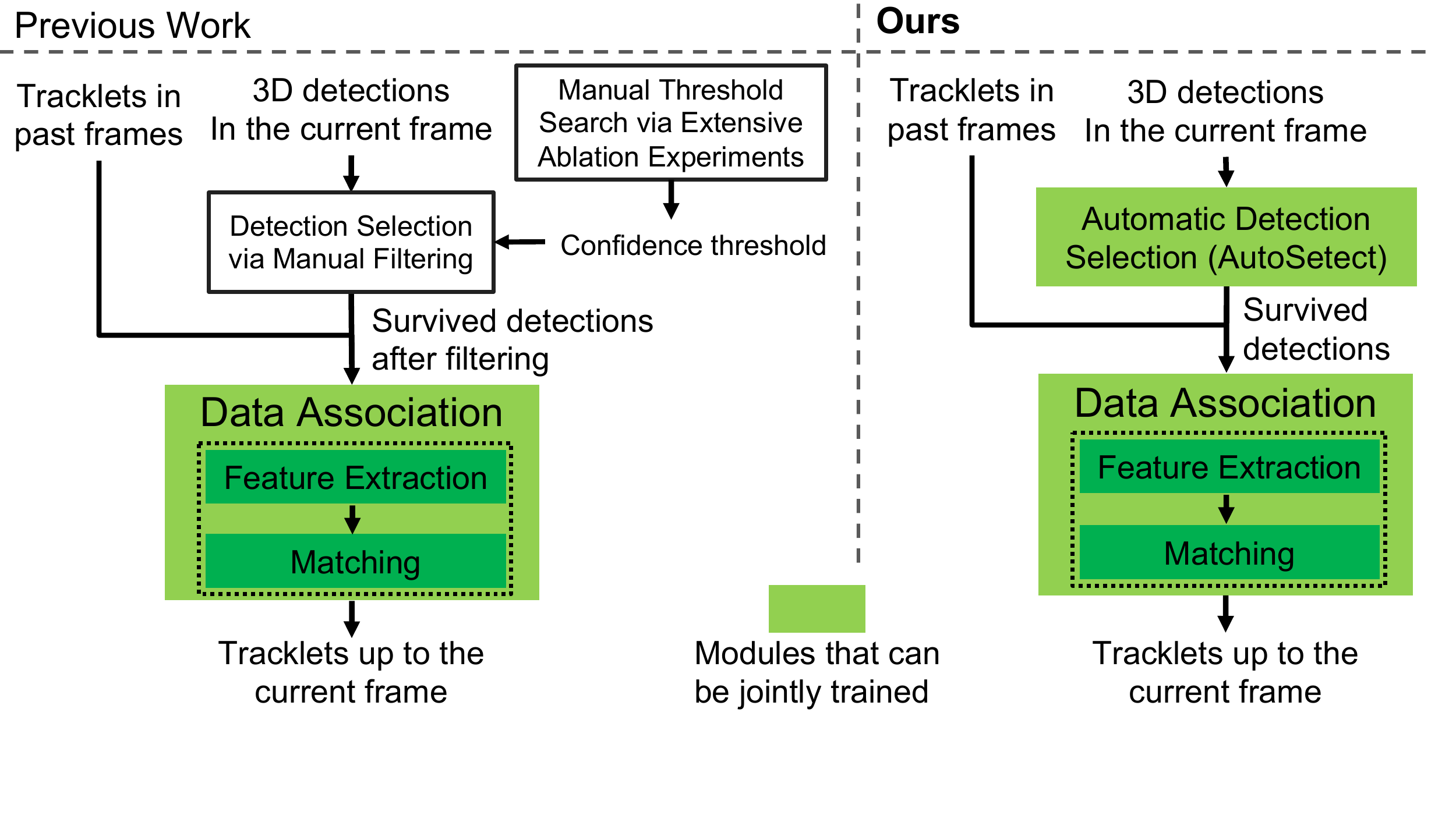}
\end{center}
  \vspace{-0.6cm}
  \caption{\textbf{Effectiveness of Automatic Detection Selection}. (Left) To improve performance, prior work needs manual threshold search to filter out false positives in pre-processing, which requires significant efforts on the user side. (Right) Our AutoSelect can automatically search for the best threshold and can be trained with data association network in an end-to-end fashion.}
  \label{fig:teaser1}
  \vspace{-0.65cm}
\end{figure}

Fortunately, a confidence score accompanied with every detection is usually provided by the 3D detector. This score indicates how confident the detector believes the detection is high-quality (\emph{i.e.,} estimated location/size are precise and predicted object category is accurate). As this score indicating detection quality is approximately close to as indicating probability of the detection being a true positive, prior work \cite{B2016, Yu2016, Wojke2017, Yuan2017, Matilla2016, B2016_2, Le2016, Weng2020_AB3DMOT, Sharma2018, Wang2020} has widely used this confidence score for detection selection (Fig. \ref{fig:teaser1} left), \emph{i.e.}, filter out detections with low scores, and has shown performance improvements. 

\begin{figure}[t]
\begin{center}
\vspace{0.15cm}
\includegraphics[trim=0.1cm 5cm 10cm 0cm, clip=true, width=\linewidth]{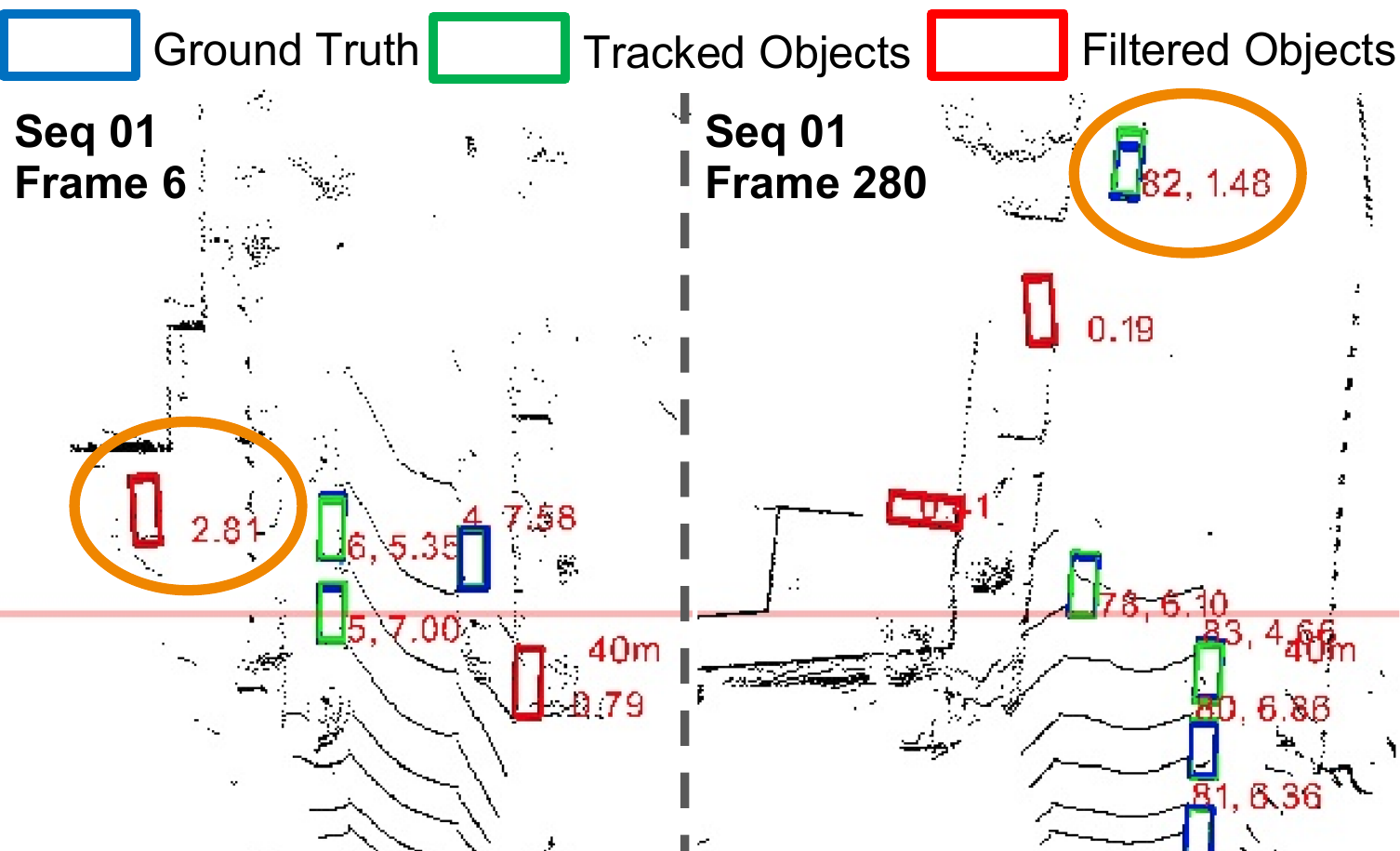}
\end{center}
  \vspace{-0.55cm}
  \caption{\textbf{Effectiveness of Dynamic Detection Selection}. Using a single threshold for detection selection might not be optimal for different frames. As highlighted in the two orange ellipses, our dynamic detection selection learns to allow true positives with low scores for data association and constructing tracklets (\emph{e.g.}, the tracked object with a score of 1.48 and a ID of 82 in the right) while filter out false positives with high scores (\emph{e.g.}, the filtered object with a score of 2.81 in the left).}
  \label{fig:teaser2}
  \vspace{-0.6cm}
\end{figure}

Despite being successful, score-based filtering for detection selection used in prior work has two main drawbacks: (1) A proper confidence score threshold needs to be manually searched, which requires non-trivial efforts on the user side. This process is usually an exhaustive/half-interval search via extensive experiments (Fig. \ref{fig:teaser1} left), \emph{i.e.}, evaluate performance on the validation set with the use of a list of thresholds, and keep the threshold achieving the highest performance. Also, this search process is usually not done once and for all. This is because the threshold is very sensitive to many factors such as data sequence, target object category and distribution of detection scores so the threshold needs to be re-searched if one of the factors changes, \emph{e.g.}, using detections produced by a different detector, switching from tracking vehicles to people; (2) As the threshold searching process is performed manually, it is impractical to adapt the threshold to every frame or every detection. As a result, prior works \cite{Weng2020_GNNTrkForecast_eccvw, Wang2020, Weng2020_AB3DMOT_eccvw, Sharma2018, Weng2020_GNN3DMOT_eccvw, Wojke2017} used a single global threshold for filtering on the entire dataset. To improve performance, a few works \cite{Jiang2019} attempted to search threshold specific to each sequence, as detections in each sequence can have different score distribution. However, using a threshold per sequence is still not optimal as score distribution can vary across frames within the same sequence, \emph{e.g.}, true positive detections might have lower scores at certain frames due to large distance to the sensor (Fig. \ref{fig:teaser2} right) or occlusion. 

To deal with the above two drawbacks of manual detection selection, we propose an automatic and dynamic detection selection technique for 3D MOT, which we call \methodnamenosp. \methodname has three advantages: (1) it can automatically select true positive detections for data association, without the need for manual threshold search via extensive experiments (Fig. \ref{fig:teaser1} right); (2) \methodname can be trained end-to-end with the data association network and optimized for the MOT objective, which is different from manual detection selection performed as a pre-processing step to data association; (3) \methodname makes it possible to perform dynamic threshold search which can lead to better detection selection in particular frames or to specific detections as shown in Fig. \ref{fig:teaser2}. To validate \methodnamenosp, we embed it into a S.O.T.A. 3D MOT method \cite{Weng2020_GNN3DMOT} and jointly train the entire system end-to-end. Unsurprisingly, as shown in the experiments, our system with \methodname is able to filter out a large portion of false positives (\emph{e.g.}, $45.7\%$ on KITTI) while preserving recall, which leads to new S.O.T.A. 3D MOT performance. To summarize, our contributions are as follows:

\begin{enumerate}
    \item Our AutoSelect removes the efforts needed for manual search of the confidence threshold in 3D MOT; 
    \item AutoSelect can dynamically search the best threshold suitable to every frame or every detection;
    \item AutoSelect for the first time makes it possible to jointly optimize detection selection with data association;
    \item Our system with automatic and dynamic detection selection achieves new S.O.T.A. 3D MOT performance. 
\end{enumerate}

\section{Related Work}

Given 2D and/or 3D sensor data, 3D MOT aims to track multiple objects in 3D space represented as 3D boxes. Recent work approaches 3D MOT in an online fashion using a tracking-by-detection pipeline \cite{Bewley2016, Weng2020_AB3DMOT}. First, an 3D object detector is applied to the current frame to obtain detections. Then, we extract discriminative features for detections in the current frame and tracklets constructed in past frames to compute a similarity matrix where each entry represents the similarity score between a pair of detection/tracklet. This matrix is fed to a bipartite matching solver such as Hungarian algorithm \cite{WKuhn1955} which assigns identities to detections and construct new tracklets in the current frame. The key factors to performance of this pipeline are two-fold: (1) object detection quality and (2) discriminative feature learning.

\subsection{Improving Detection Quality in 3D MOT}

The most important factor to performance of the tracking-by-detection pipeline is detection quality, \emph{i.e.}, false positives (FPs) and false negatives (FNs). This is because, if a target cannot be detected (\emph{i.e.}, a FN) for a few frames, it is hard to match past tracklets with the target and construct tracklets of the target in following frames. Also, if there are many FP detections that continue to appear for a few frames, it is easy to construct FP tracklets for non-existing objects. Both such cases will degrade tracking performance.

Historically, researchers improve detection quality by developing better detectors. To that end, a large number of 3D object detectors have been proposed to improve performance. For example, for the car class on the KITTI \cite{Geiger2012} 3D detection benchmark, AP (average precision) in the moderate level of difficulty has improved from 66.47 \cite{Ku2018} to 81.43 \cite{Shi2020} in two years. To achieve S.O.T.A performance, prior work often leverages point clouds captured by the LiDAR sensor, which provide accurate depth information. The differences in prior work lie in how to process point clouds, which include (1) methods \cite{Beltran2018, Ku2018, Xiaozhi2017} that project point clouds into bird's eye view images and apply 2D Convolutional Neural Networks (CNNs), (2) approaches \cite{Luo2018, Yan2018, Zhou2018} that voxelize clouds into 3D tensors and then use 3D CNNs, or (3) methods that process point clouds with point-based convolution \cite{Shi2020, Qi2018, Shi2019}. However, given a fixed 3D object detector (\emph{e.g.}, the best available detector at the moment), how could we further improve detection quality?

One solution to the above question is to pre-process the detections obtained from the best detector. For example, we can reduce FPs by score-based filtering. Also, we can reduce FNs by interpolating tracklet segments corresponding to the same object. Although pre-processing techniques have been widely used in prior work \cite{B2016, Yu2016, Wojke2017, Yuan2017, Matilla2016, B2016_2, Le2016, Weng2020_AB3DMOT, Sharma2018, Wang2020, Jiang2019} to improve performance, they are often based on heuristics and require a significant amount of efforts on the user side to manually search for a proper threshold. Different from prior work, we for the first time propose an automatic technique to improve detection quality given fixed detections, with a specific focus on detection selection, \emph{i.e.}, reducing the number of FPs. Also, our automatic detection selection is no longer a pre-processing step but can be integrated into any modern 3D MOT systems with end-to-end training.

\begin{figure*}[t]
\begin{center}
\includegraphics[trim=0.1cm 1.8cm 0cm 0cm, clip=true, width=0.87\linewidth]{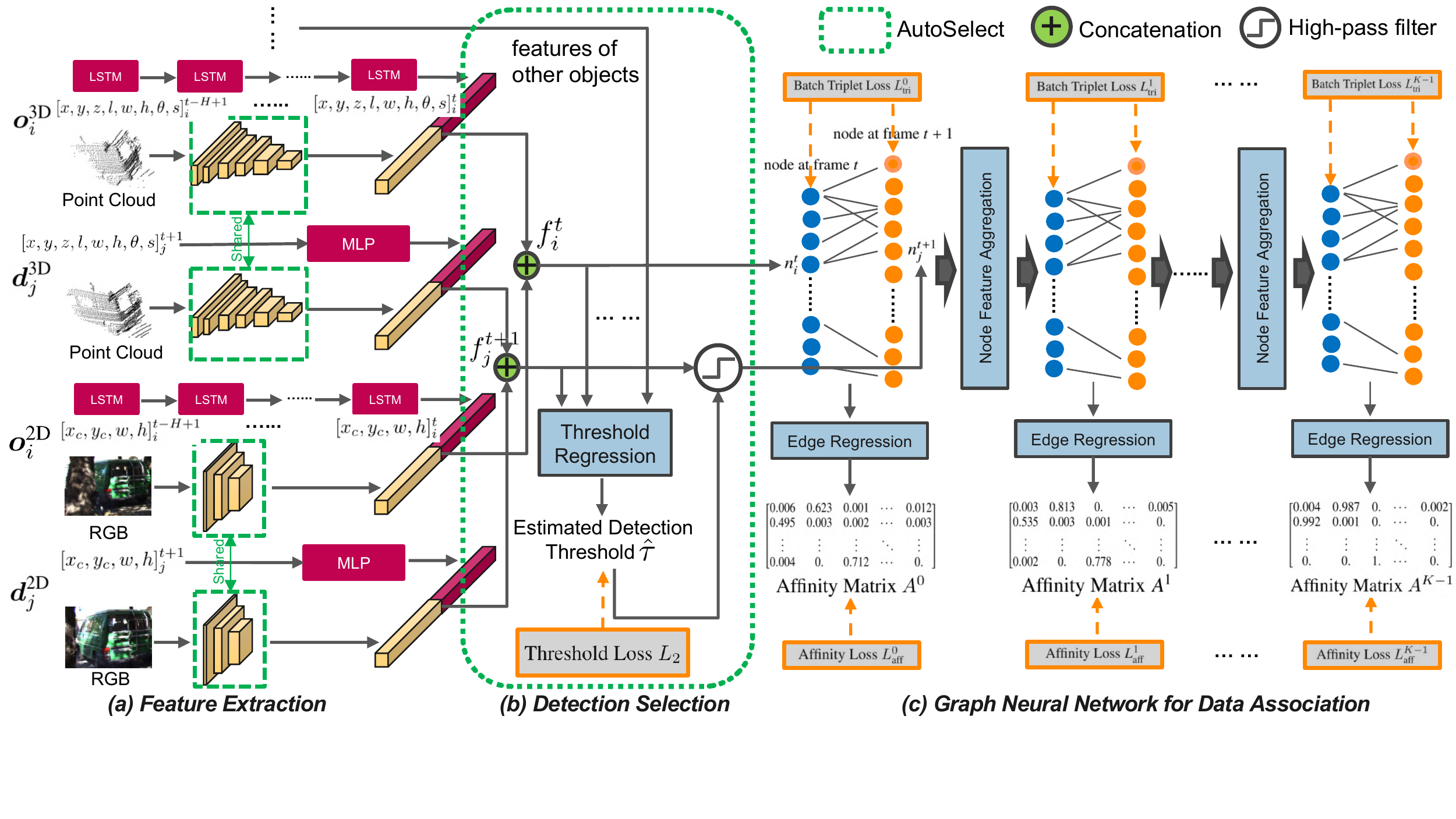}
\end{center}
  \vspace{-0.6cm}
  \caption{\textbf{Proposed System:} (a) A feature extraction module is used to learns multi-modal features for each tracked object $\boldsymbol{o}_i$ and for each 3D detection $\boldsymbol{d}_j$; (b) A detection selection module is used to automatically and dynamically estimate detection threshold every frame and filter out false positive detections with scores lower than the threshold; (c) A Graph Neural Network is used to encode contextual features between survived objects by modeling interaction.}
  \label{fig:main}
  \vspace{-0.6cm}
\end{figure*}

\subsection{Discriminative Feature Learning in 3D MOT}

To obtain discriminative feature, prior work has explored different types of features, among which motion \cite{Weng2020_AB3DMOT, Bewley2016, Caselitz2016} and appearance features \cite{Sun2017, Frossard2018, Yu2016, Zhang2019_robust} turn out to be the most discriminative. To leverage complementary information in both features, recent work \cite{Baser2019, Weng2020_GNN3DMOT} tends to fuse two features. Also, beyond learning isolated features for each object, recent work \cite{Weng2020_GNN3DMOT, Weng2020_GNNTrkForecast, Braso2020} improves discriminative feature learning by encoding contextual features and modeling interaction using Graph Neural Networks (GNNs), which has largely improved performance. To leverage the success in the literature, we build our system on top of \cite{Weng2020_GNN3DMOT} which has both advantages of feature fusion and GNN-based interaction modeling. By embedding AutoSelect into \cite{Weng2020_GNN3DMOT}, our system has both the capabilities of improving detection quality and learning discriminative features.

\section{Approach}

Let $\mathcal{O}$=$\{\boldsymbol{o}_1, \ldots, \boldsymbol{o}_M\}$ denotes the set of $M$ tracklets in past frames. Each tracklet $\boldsymbol{o}_i$=$[\boldsymbol{o}_i^{-H}, \ldots, \boldsymbol{o}_i^{-1}]$ consists of associated 3D detections of the $i$-th tracked object in the past $H$ frames. The associated 3D detection at frame $\xi \in \{t$-$H$+$1$, $\ldots$, $t$-$1$, $t\}$ is a tuple $\boldsymbol{o}_{i,\xi}^{\text{3D}}$=[$x$, $y$, $z$, $l$, $w$, $h$, $\theta$, $s$, $I$]${}_i^\xi$, where ($x, y, z$) is the object center in 3D space, $(l, w, h)$ is the object size, $\theta$ is the heading angle, $s$ is the confidence score, and $I$ is the assigned ID. Let $\mathcal{D}$=$\{\boldsymbol{d}_1, \ldots, \boldsymbol{d}_N\}$ denotes the set of $N$ 3D detections in the current frame $t$+$1$ by a 3D detector. Each 3D detection $\boldsymbol{d}_j^{\text{3D}}$=$[x, y, z, l, w, h, s, \theta]_{j}^{t+1}$ is defined similarly to the past associated detections $\boldsymbol{o}_i^t$ except without the assigned ID $I$. The goal of online 3D MOT is to associate 3D detection $\boldsymbol{d}_j \in \mathcal{D}$ with tracklet $\boldsymbol{o}_i \in \mathcal{O}$ and assign an ID to $\boldsymbol{d}_j$, and then perform this data association process over the entire sequence. 

We illustrate our 3D MOT system in Fig. \ref{fig:main}, which contains three modules: (a) a feature extraction module that extracts multi-modal discriminative features (2D motion, 3D motion, 2D appearance, 3D appearance) from $O$ and $D$; (b) an automatic and dynamic detection selection module that estimates a detection threshold $\hat{\tau}$ per frame and filter out detections with scores lower than $\hat{\tau}$; (c) a GNN that takes the features of non-filtered objects as nodes and encodes contextual features for each object node via interaction. 

\subsection{Feature Extraction}

To extract complementary features from different modalities, our feature extraction has four branches for 3D motion/appearance and 2D motion/appearance respectively.

\vspace{1.5mm}\noindent\textbf{3D Motion Feature.} For each tracklet $\boldsymbol{o}_i$, we use a two-layer LSTM \cite{Hochreiter1997} with a hidden dimension of $32$ to extract 3D motion feature. The input to the LSTM at every time stamp is the 3D box $\boldsymbol{o}_{i,\xi}^{\text{3D}}$=[$x$, $y$, $z$, $l$, $w$, $h$, $\theta$, $s$]${}_i^\xi$ (no need to include $I$ here). We use past $H$=$5$ frames to obtain information from history. For each 3D detection $\boldsymbol{d}_j$, we only have its 3D box in the current frame $\boldsymbol{d}_j^{\text{3D}}$=$[x, y, z, l, w, h, s, \theta]_{j}^{t+1}$ (not a tracklet), we use a two-layer (8$\Rightarrow$16$\Rightarrow$32) MLP (Multi-Layer Perceptron) to extract its location information. 

\vspace{1.5mm}\noindent\textbf{3D Appearance Feature.} For each $\boldsymbol{o}_i$ or $\boldsymbol{d}_j$, we crop the point cloud enclosed by its 3D box with a size of $P$ (number of points) $\times$ 4 ($x$, $y$, $z$, reflectance). Then, we feed the point cloud to a PointNet \cite{Cherabier2017} (4$\Rightarrow$16$\Rightarrow$32$\Rightarrow$64$\Rightarrow$128 $\Rightarrow$64$\Rightarrow$32) to extract the 3D appearance feature. As there is a list of clouds for tracklets (one cloud per time stamp), it will be computationally expensive to process all these clouds, so we only use the cloud in the most recent frame $t$ for tracklets. The PointNet is shared for all tracklets $\mathcal{O}$ and detections $\mathcal{D}$.

\vspace{1.5mm}\noindent\textbf{2D Motion Feature.} In addition to 3D features, we leverage 2D features. To obtain 2D box corresponding to each 3D box, we project eight corners of the 3D box into image plane given camera matrix and compute the minimum bounding rectangle of the projected corners as the 2D box. Formally, each 2D box is parameterized as $\boldsymbol{d}_j^{2D}$=$[x_c, y_c, w, h]_j^{t+1}$ or $\boldsymbol{o}_{i,\xi}^{2D}$=$[x_c, y_c, w, h]_i^\xi$ where ($x_c, y_c$) is the object center in image plane, ($w$, $h$) is width and height of the 2D box. Since the confidence score $s$ has been already encoded in 3D features, we do not add $s$ into the 2D parameterization. Similar to 3D motion feature extraction, we feed tracklets $\boldsymbol{o}_{i}^{2D}$ to a two-layer LSTM with a dimension of $32$ to obtain 2D motion feature. For 2D detection $\boldsymbol{d}_j^{2D}$, we use a two-layer MLP (4$\Rightarrow$8$\Rightarrow$32) to extract its 2D location information.

\vspace{1.5mm}\noindent\textbf{2D Appearance Feature.} We use image crop to extract 2D appearance feature. Given 2D box of $\boldsymbol{d}_j^{2D}$ or $\boldsymbol{o}_{i}^{2D}$, we crop its image patch and feed it to a ResNet-34~\cite{He2016} followed by a fully connected layer to obtain 2D appearance feature with a size of $32$. Similar to the 3D appearance branch, we only use the most recent frame of image crop for 2D tracklets.

We apply the above feature extraction to all objects in $\mathcal{O}$ and $\mathcal{D}$, though we only illustrate for one tracklet $\boldsymbol{o}_i$ and one detection $\boldsymbol{d}_j$ in Fig. \ref{fig:main}. Then, we fuse the extracted 2D/3D motion/appearance features by concatenation to obtain the final features with a dimension of $128$ (\emph{e.g.}, $f_j^{t+1}$ for $\boldsymbol{d}_j$ and $f_i^t$ for $\boldsymbol{o}_i$), which are then used in the following modules. 


\subsection{Automatic and Dynamic Detection Selection}

To filter out FPs given detections and improve quality of detections used for data association, the key is to perform detection selection as shown in Fig. \ref{fig:main} (b). Our detection selection module takes output features from the feature extraction module and estimates a detection threshold $\hat{\tau}$ every frame via a threshold regression module. Then, we pass the estimated threshold and all raw detections as inputs to a high-pass filter which only selects feature of objects that have a raw score higher than the threshold $\hat{\tau}$.

\subsubsection{Threshold Regression Network (TRN)}

To estimate the detection threshold $\hat{\tau}$ (a single scalar) from high-dimensional features, it is straightforward to use a three-layer MLP (128$\Rightarrow$32 $\Rightarrow$8$\Rightarrow$1). As the number of objects including tracklets and detections can vary across frames, the input to TRN is a variable-sized feature with the size of $(M+N)\times128$. As a result, a max-pooling (MP) operator is needed to reduce the size of the feature to be fixed which is required by MLP. 

In addition to investigate the architecture of TRN, another important question is what information we should feed into TRN to achieve the best detection selection. As information of tracked objects in past frames such as their locations and sizes could be useful to identity FPs in the current frame, one straightforward option is to feed high-dimensional features of $\mathcal{O}$ and $\mathcal{D}$ all together into TRN as shown in Fig. \ref{fig:main}. To validate if information of past objects is actually useful to detection selection, an alternative is to use only features of detections as inputs to TRN. Also, as data association and detection selection might require very different features, using the same features for both tasks as shown in Fig. \ref{fig:main} might not be optimal. So we experimented using raw 3D boxes including scores of detections or of detections/tracklets as inputs to TRN in the ablation study. 

To train our TRN, it is important to obtain a proper detection threshold $\tau$ as ground truth. Given a list of detections including their scores $\mathcal{S}=\{s_1, s_2, \cdots, s_N\}$ and ground truth (GT) object boxes in the current frame, we first match the detections with GT using the Hungarian algorithm to identity which detections are FPs and which are true positives (TP). Also, an Intersection of Union (IoU) threshold of $\text{IoU}_{\text{min}}$ is used for the matching. Once we are aware of the attribute of detections (FP or TP), we can split $\mathcal{S}$ into $\mathcal{S}_{\text{TP}}$ and $\mathcal{S}_{\text{FP}}$ where they denote a list of scores of detections which are TPs and FPs respectively. As the goal of AutoSelect is to filter out most FPs while preserve TPs and recall, the most straightforward solution is to obtain GT $\tau$ is as follows:
\vspace{-0.1cm}
\begin{equation}
   \tau = \min ( \mathcal{S}_{\text{TP}}) - s_{\text{buff}},
   \vspace{-0.1cm}
   \label{eq:gt_1}
\end{equation}
where $s_{\text{buff}}$ is a scalar that gives some buffer space between the threshold boundary and minimum score of all TPs. In practice, we found that using $s_{\text{buff}}=3$ works quite well. In fact, if the estimated threshold $\hat{\tau}$ can be perfectly close to GT $\tau$, there will be no TP filtered out (perfectly maintain the maximum recall) while many FPs can be filtered out. However, one disadvantage of Eq. \ref{eq:gt_1} is that it can cause instability during TRN training as there is no constraint of upper bound of $\tau$. For example, sometimes there is only one very confident TP detection that is very close and has a high score (\emph{e.g.}, 13.0) and a few weak TP detections that are far with low scores (\emph{e.g.}, 3.5, 3.1). As those far objects are hard to detect, it is possible that they are detected at frame $t$ but missed at frame $t+1$. In such case, the GT $\tau$ computed by Eq. \ref{eq:gt_1} at frame $t$ and $t+1$ are very different: $10.0$ v.s $0.1$, \emph{i.e.}, suddenly become much larger in frame $t+1$. If we force TRN to learn with such GT $\tau$, the network can easily predict a very large threshold at test time which filters out many TPs and disobey our goal. In fact, we observed that this case happens very frequently so that adding an upper bound to the GT threshold as below is needed:
\vspace{-0.1cm}
\begin{equation}
   \tau = \min( \min ( \mathcal{S}_{\text{TP}}) - s_{\text{buff}}, s_{\text{upper}}),
   \vspace{-0.1cm}
   \label{eq:gt_2}
\end{equation}
where $s_{\text{upper}}$ is a scalar as the upper bound of $\tau$. By analyzing distribution of scores of FPs/TPs produced by the 3D detector \cite{Shi2019}, we found $s_{\text{upper}}=3$ statistically works well in practice. For example, in the training set of KITTI tracking dataset, $86.29\%$ of FP detections by \cite{Shi2019} have scores lower than $3$ while $89.82\%$ TP detections have scores higher than $3$. Once we have computed the GT $\tau$, we can now train the TRN via a $L2$ loss by minimizing the difference between the estimated threshold $\hat{\tau}$ and GT detection threshold $\tau$.

\begin{figure}[t]
\begin{center}
\vspace{0.15cm}
\includegraphics[trim=0.4cm 5cm 15.9cm 0.2cm, clip=true, width=0.5\linewidth]{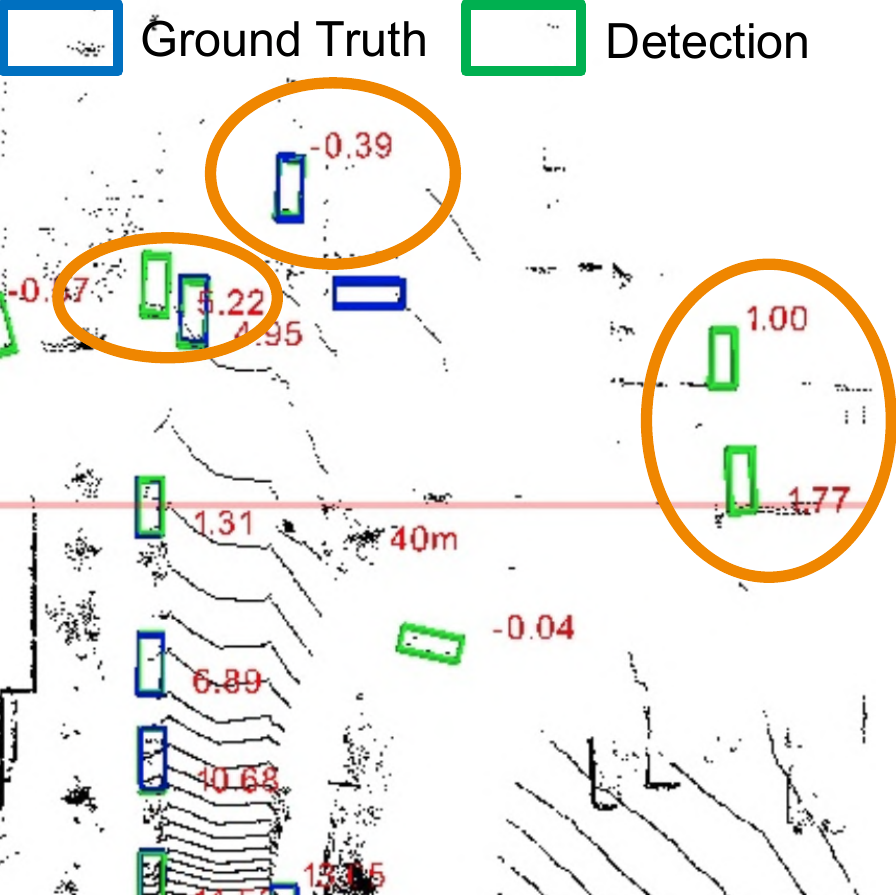}
\includegraphics[trim=0.1cm 2.1cm 13.3cm 0cm, clip=true, width=0.485\linewidth]{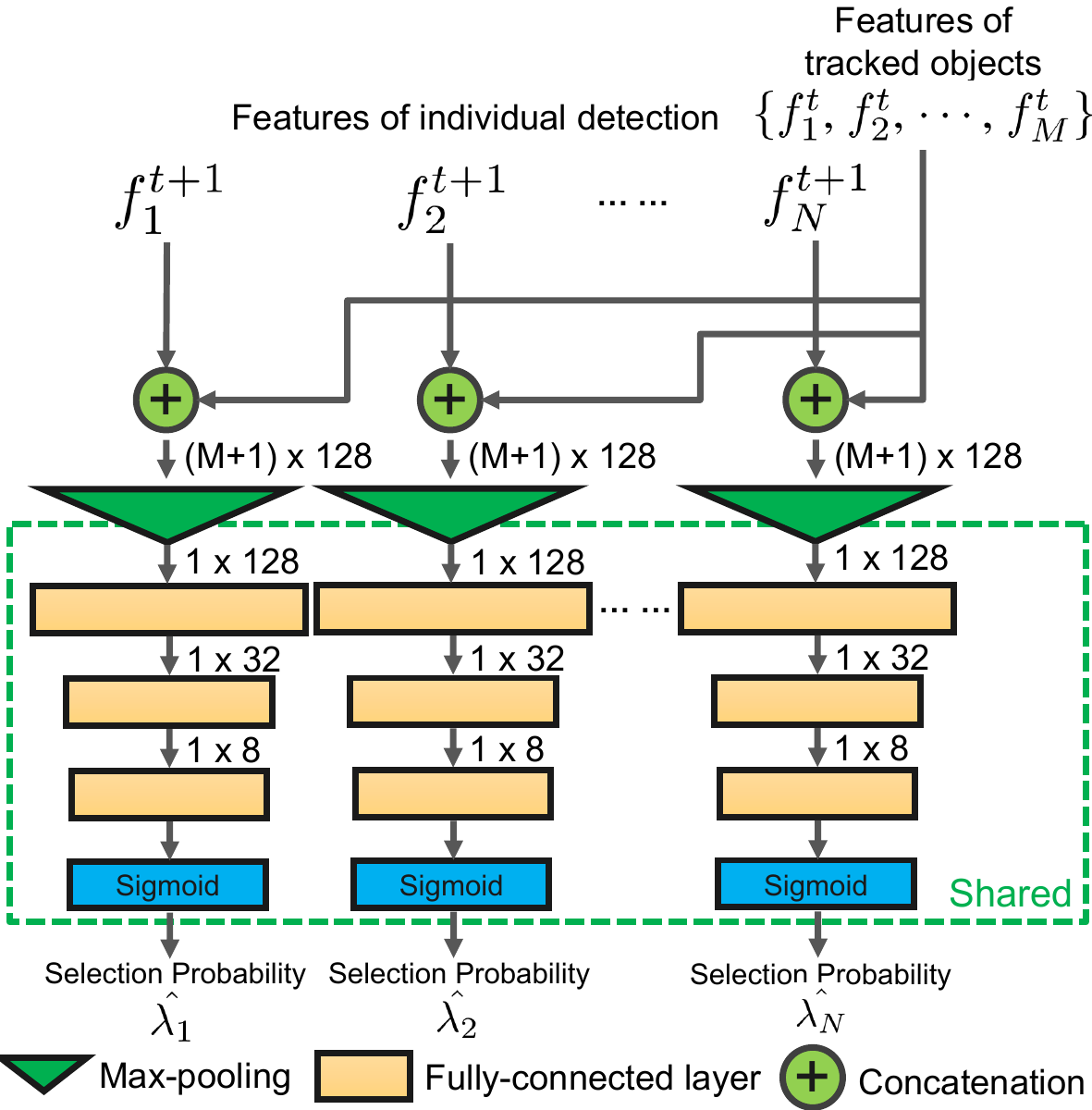}
\end{center}
  \vspace{-0.55cm}
  \caption{\textbf{(Left)} Using a per-frame threshold cannot separate TP/FP detections perfectly. As highlighted in orange ellipses, a TP detection (\emph{e.g.}, the top one with a score of -0.39) can have a lower score than other FP detections (\emph{e.g.}, the FP detections with a score of 5.22 in the left and with scores of 1.00/1.77 in the right ellipse). Note that only \textcolor{Green}{detections} have scores beside and a TP detection means that a \textcolor{Green}{detection} is overlapped with a \textcolor{blue}{GT}. \textbf{(Right)} The network architecture of our instance-level AutoSelect, which aims to estimate the probability of being a TP for each individual detection.}
  \label{fig:instance}
  \vspace{-0.5cm}
\end{figure}

\begin{table*}[t]
\caption{3D MOT performance on the car category of KITTI dataset.}
\vspace{-0.3cm}
\centering
\resizebox{\textwidth}{!}{
\begin{tabular}{@{}llrrrrrrrrr@{}}
\toprule
Evaluation matching criteria & Method & \textbf{sAMOTA}$\uparrow$ & AMOTA$\uparrow$ & AMOTP$\uparrow$ & \textbf{MOTA}$\uparrow$ & MOTP$\uparrow$ & IDS$\downarrow$ & FRAG$\downarrow$ & FP$\downarrow$ & FN$\downarrow$ \\
\midrule
3D $\text{IoU}_{\text{thres}}$ = 0.25 
& mmMOT~\cite{Zhang2019_robust} (ICCV $'$19) & 70.61 & 33.08 & 72.45 & 74.07 & 78.16 & 10 & 55 & 946 & 1012 \\
& FANTrack~\cite{Baser2019} (IV $'$20) & 82.97 & 40.03 & 75.01 & 74.30 & 75.24 & 35 & 202 & 1146 & 751 \\
& AB3DMOT~\cite{Weng2020_AB3DMOT} (IROS $'$20) & 93.28 & 45.43 & 77.41 & 86.24 & 78.43 & \textbf{0} & 15 & 365 & 788 \\
& GNN3DMOT~\cite{Weng2020_GNN3DMOT} (CVPR $'$20) & \textbf{93.68} & 45.27 & 78.10 & 84.70 & \textbf{79.03} & 23 & 54 & 470 & 795 \\
& \textbf{Frame-level AutoSelect (Ours)} & 93.46 & 45.97 & \textbf{78.20} & 87.27 & 78.86 & \textbf{0} & \textbf{9} & 324 & \textbf{743} \\
& \textbf{Instance-level AutoSelect (Ours)} & 93.66 & \textbf{46.52} & 78.05 & \textbf{87.48} & 78.99 & 3 & 10 & \textbf{255} & 791 \\
\midrule
3D $\text{IoU}_{\text{thres}}$ = 0.5 
& mmMOT~\cite{Zhang2019_robust} (ICCV $'$19) & 69.14 & 32.81 & 72.22 & 73.53 & 78.51 & 10 & 64 & 849 & 1296 \\ 
& FANTrack~\cite{Baser2019} (IV $'$20) & 80.14 & 38.16 & 73.62 & 72.71 & 74.91 & 36 & 211 & 1263 & 915 \\
& AB3DMOT~\cite{Weng2020_AB3DMOT} (IROS $'$20) & 90.38 & 42.79 & 75.65 & 84.02 & 78.97 & \textbf{0} & 51 & 480 & 859 \\
& GNN3DMOT~\cite{Weng2020_GNN3DMOT} (CVPR $'$20) & 90.24 & 42.73 & 76.38 & 81.17 & \textbf{79.86} & 41 & 113 & 475 & 1062 \\
& \textbf{Frame-level AutoSelect (Ours)} & 90.60 & 43.24 & 76.40 & 85.13 & 79.70 & \textbf{0} & \textbf{49} & 340 & 906 \\
& \textbf{Instance-level AutoSelect (Ours)} & \textbf{92.85} & \textbf{45.73} & \textbf{78.31} & \textbf{85.48} & 79.55 & 2 & 59 & \textbf{304} & \textbf{903} \\
\midrule
3D $\text{IoU}_{\text{thres}}$ = 0.7 
& mmMOT~\cite{Zhang2019_robust} (ICCV $'$19) & 63.91 & 24.91 & 67.32 & 51.91 & 80.71 & 24 & 141 & 980 & 2970 \\ 
& FANTrack~\cite{Baser2019} (IV $'$20) & 62.72 & 24.71 & 66.06 & 49.19 & 79.01 & 38 & 406 & 1328 & 2871 \\
& AB3DMOT~\cite{Weng2020_AB3DMOT} (IROS $'$20) & 69.81 & 27.26 & 67.00 & 57.06 & 82.43 & \textbf{0} & 157 & 966 & 2632 \\
& GNN3DMOT~\cite{Weng2020_GNN3DMOT} (CVPR $'$20) & 73.22 & 29.09 & 67.22 & 61.62 & 82.57 & 5 & 214 & 787 & 2424 \\
& \textbf{Frame-level AutoSelect (Ours)} & 76.01 & 31.40 & 69.29 & 63.48 & 82.50 & \textbf{0} & 209 & 668 & 2436 \\
& \textbf{Instance-level AutoSelect (Ours)} & \textbf{76.13} & \textbf{31.41} & \textbf{69.33} & \textbf{65.46} & \textbf{82.61} & \textbf{0} & \textbf{192} & \textbf{515} & \textbf{2421} \\
\bottomrule
\end{tabular}}
\vspace{-0.75cm}
\label{tab:kitti_quan}
\end{table*}

\subsubsection{Frame-level vs. Instance-level}

With the above TRN, we can dynamically adapt detection selection to every frame. However, would it be possible that in the same frame the best detection selection rule is different for each detection? Our answer is yes. For example in Fig. \ref{fig:instance} (left), in the same frame, a TP detection might have a lower score than FP detections due to the fact this TP object is far from the sensor or is occluded by objects in front so its detection is highly uncertain with a low score. In this case, frame-level AutoSelect cannot perfectly filter out the FP while preserve the TP using a per-frame threshold, \emph{i.e.}, it will either filter out all or preserve both the TP/FP detections. 

To deal with the issue that a per-frame detection threshold cannot perfectly separate TPs and FPs (\emph{i.e.}, filter out all FPs and preserve all TPs), we propose a variant of AutoSelect for instance-level detection selection as shown in Fig. \ref{fig:instance} (right). Instance-level AutoSelect aims to estimate the probability of being a TP (\emph{i.e.} true positiveness) for each detection at every frame. Specifically, for each detection $\boldsymbol{d}_j$, we use its own feature $f_j^{t+1}$ and features of tracked objects $\{f_1^t, f_2^t, \cdots, f_M^t\}$ as inputs to a MLP+MP network (the same as TRN) with an additional Sigmoid function in the end to obtain its selection probability $\hat{\lambda}_j$. Also, this network is shared for all detections.

To train our instance-level AutoSelect, we need GT for selection probability, which is however very simple to obtain in the instance-level case. After we identify TPs/FPs by matching detections with GT objects, we assign TP detections with GT $\lambda$ of $1$ and FP detections with GT $\lambda$ of $0$. As the instance-level AutoSelect is formulated as a classification problem, we use a binary-cross entropy loss to train the network. At test time, we use a probability threshold $\lambda_{\text{thres}} = 0.1$ to select detections if their $\hat{\lambda}$ are larger than $\lambda_{\text{thres}}$. Note that our selection probability $\hat{\lambda}$ is fundamentally different from the confidence score. This is because our selection probability $\hat{\lambda}$ denotes true positiveness and can better separate TP/FP detections while the confidence score used in prior work is only an approximation of true positiveness and cannot well separate TP/FP detections as shown in Fig. \ref{fig:instance} (left).

\subsection{Graph Neural Networks for Data Association}

Following \cite{Weng2020_GNN3DMOT}, we use GNNs to model object interactions and encode contextual features. Similarly, we use extracted features $f$ of survived objects (not including objects filtered out) as nodes to construct a graph. Then, GNNs are used to update features via node feature aggregation. As we use the same aggregation rule as \cite{Weng2020_GNN3DMOT}, we refer readers to \cite{Weng2020_GNN3DMOT} for details. At each layer, we estimate an affinity matrix $A$ by feeding node features to a MLP for edge regression as shown in Fig. \ref{fig:main}. To train our data association network, we apply an affinity loss $L_{\mathrm{aff}}$ to the estimated affinity matrix $A$ and apply a batch triplet loss $L_{\mathrm{tri}}$ to the node features, at every layer of the GNNs. Please refer to \cite{Weng2020_GNN3DMOT} for detailed loss equations. We use the same weight of $1$ for all three losses (including loss for AutoSelect). At test time, we feed the estimated affinity matrix $A$ to the Hungarian algorithm to obtain the matching. Then, we use the same tracking management as \cite{Weng2020_GNN3DMOT} for tracklet initialization and termination.

\begin{figure*}[t]
\begin{center}
\includegraphics[trim=2.5cm 3.2cm 2.5cm 5.3cm, clip=true, width=0.326\linewidth]{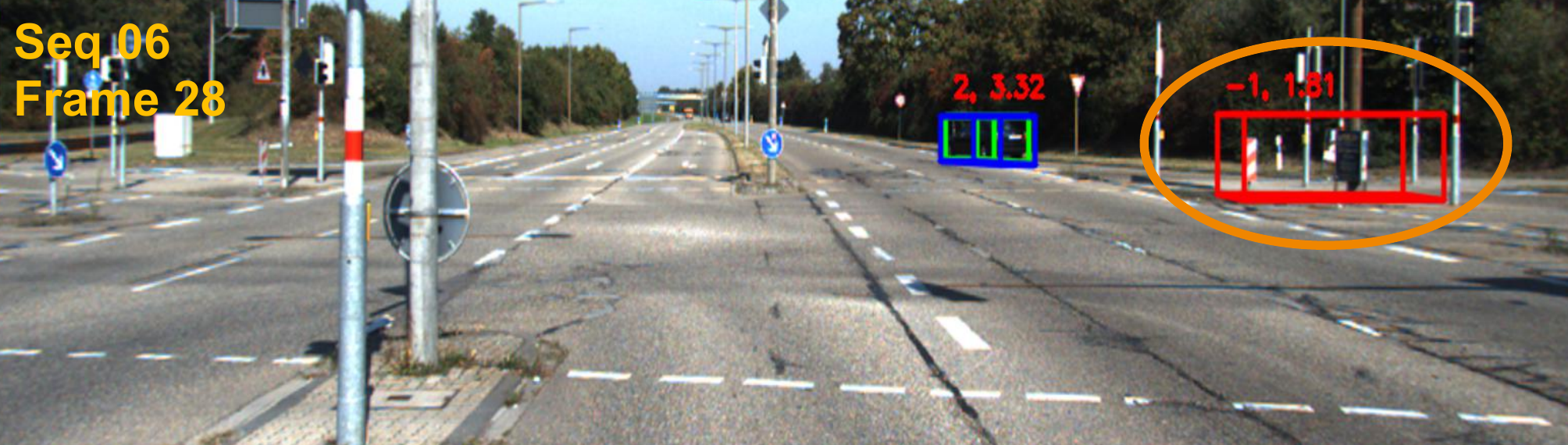}
\includegraphics[trim=2.5cm 3.2cm 2.5cm 5.3cm, clip=true, width=0.326\linewidth]{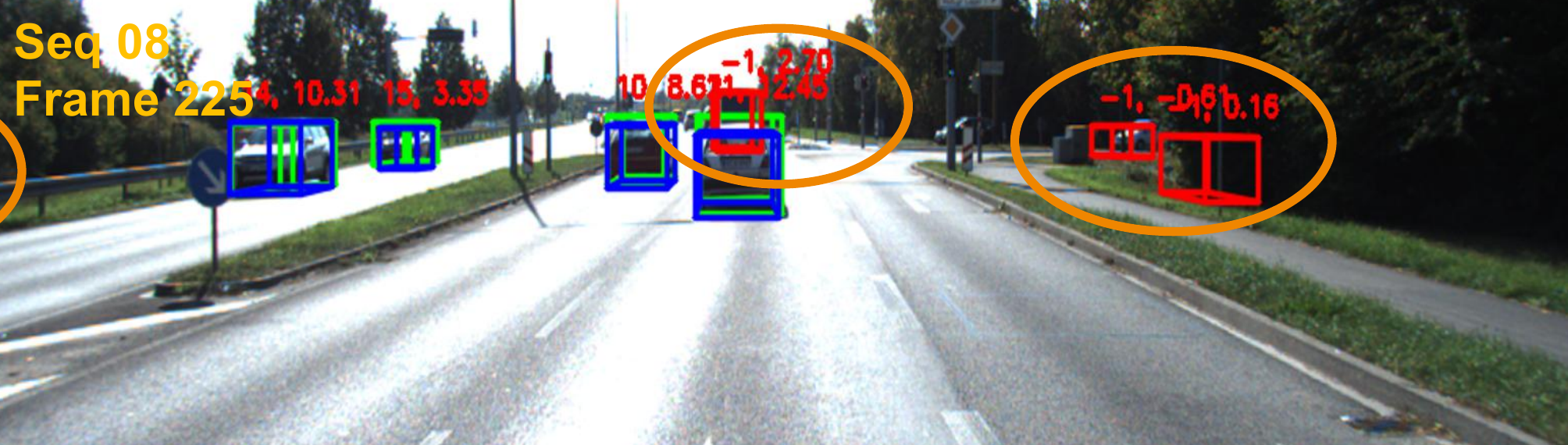}
\includegraphics[trim=4.9cm 3.2cm 0.1cm 5.3cm, clip=true, width=0.326\linewidth]{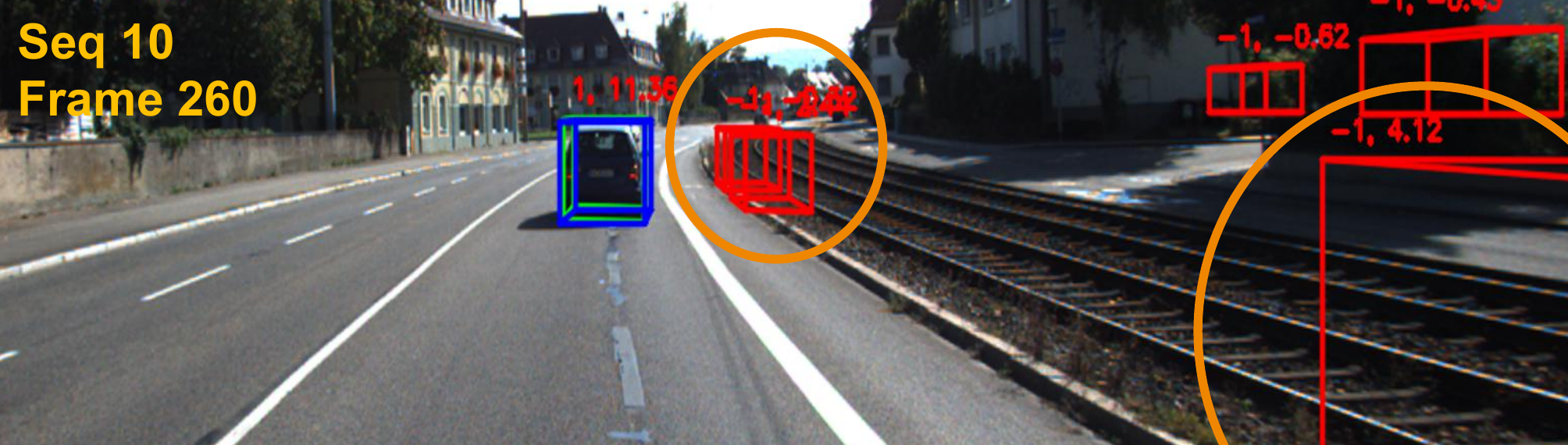}
\includegraphics[trim=0.1cm 2.5cm 10.5cm 0.8cm, clip=true, width=0.326\linewidth]{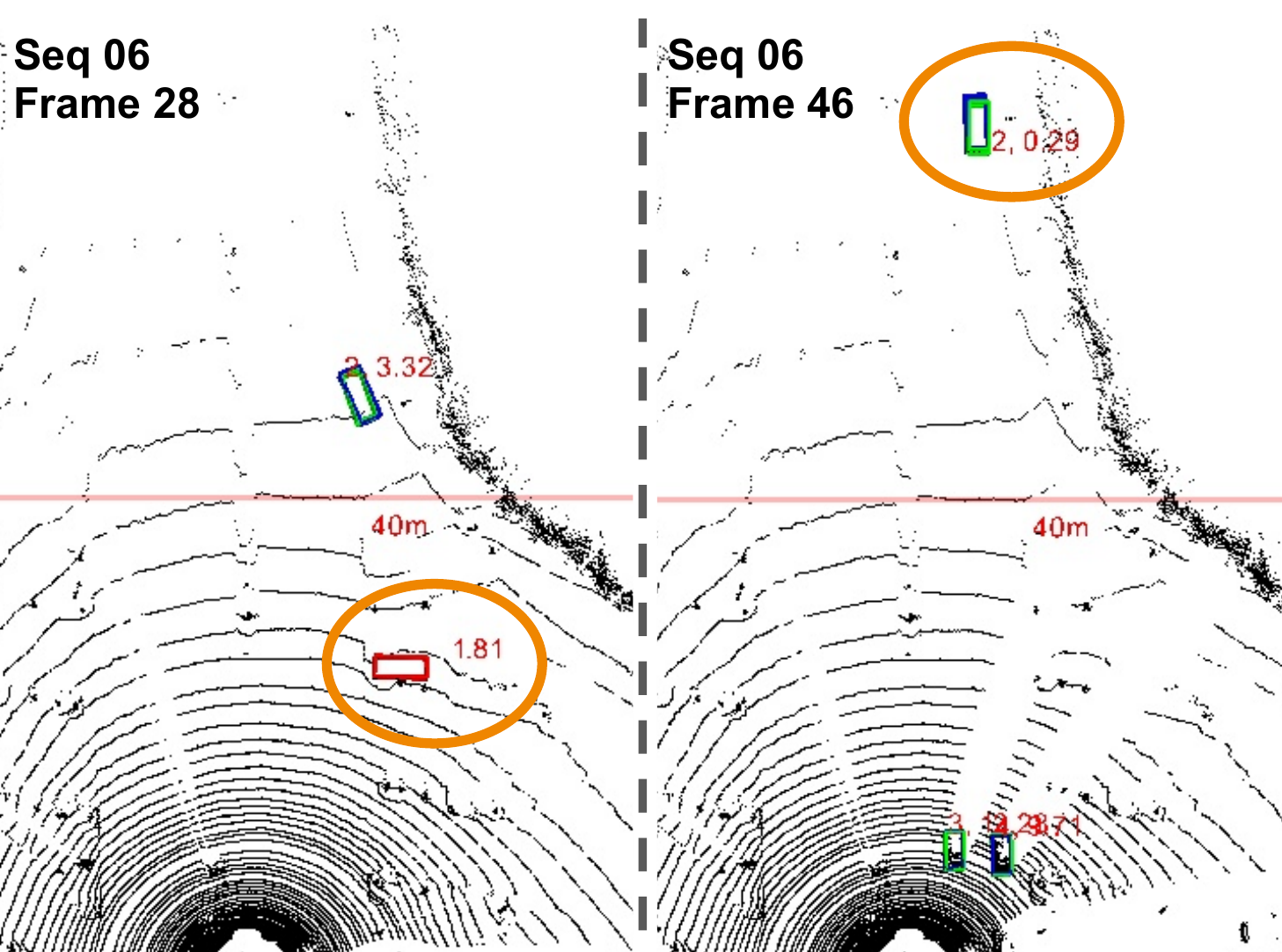}
\includegraphics[trim=0.1cm 2.5cm 10.5cm 0.8cm, clip=true, width=0.326\linewidth]{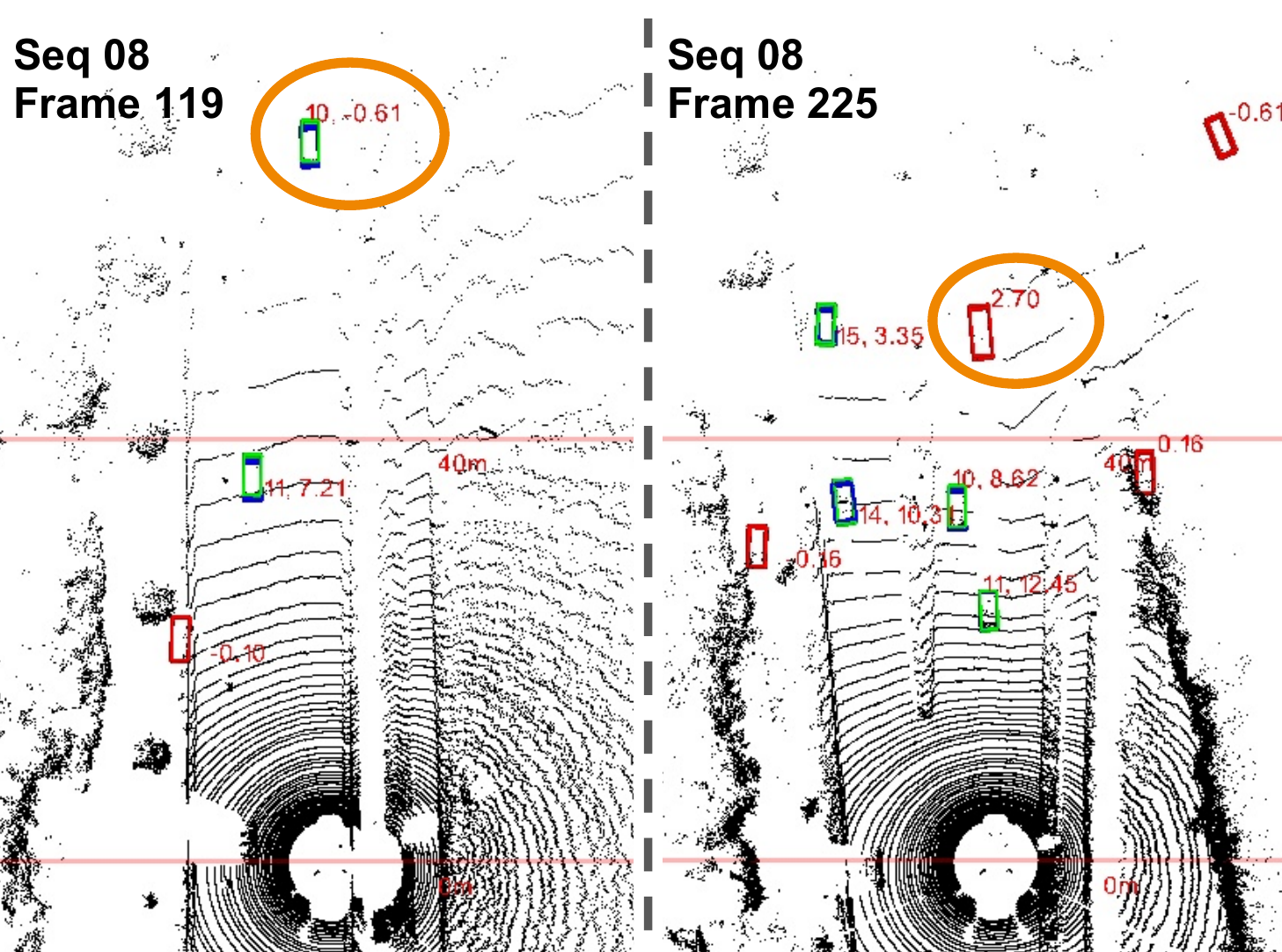}
\includegraphics[trim=0.1cm 2.5cm 10.5cm 0.8cm, clip=true, width=0.326\linewidth]{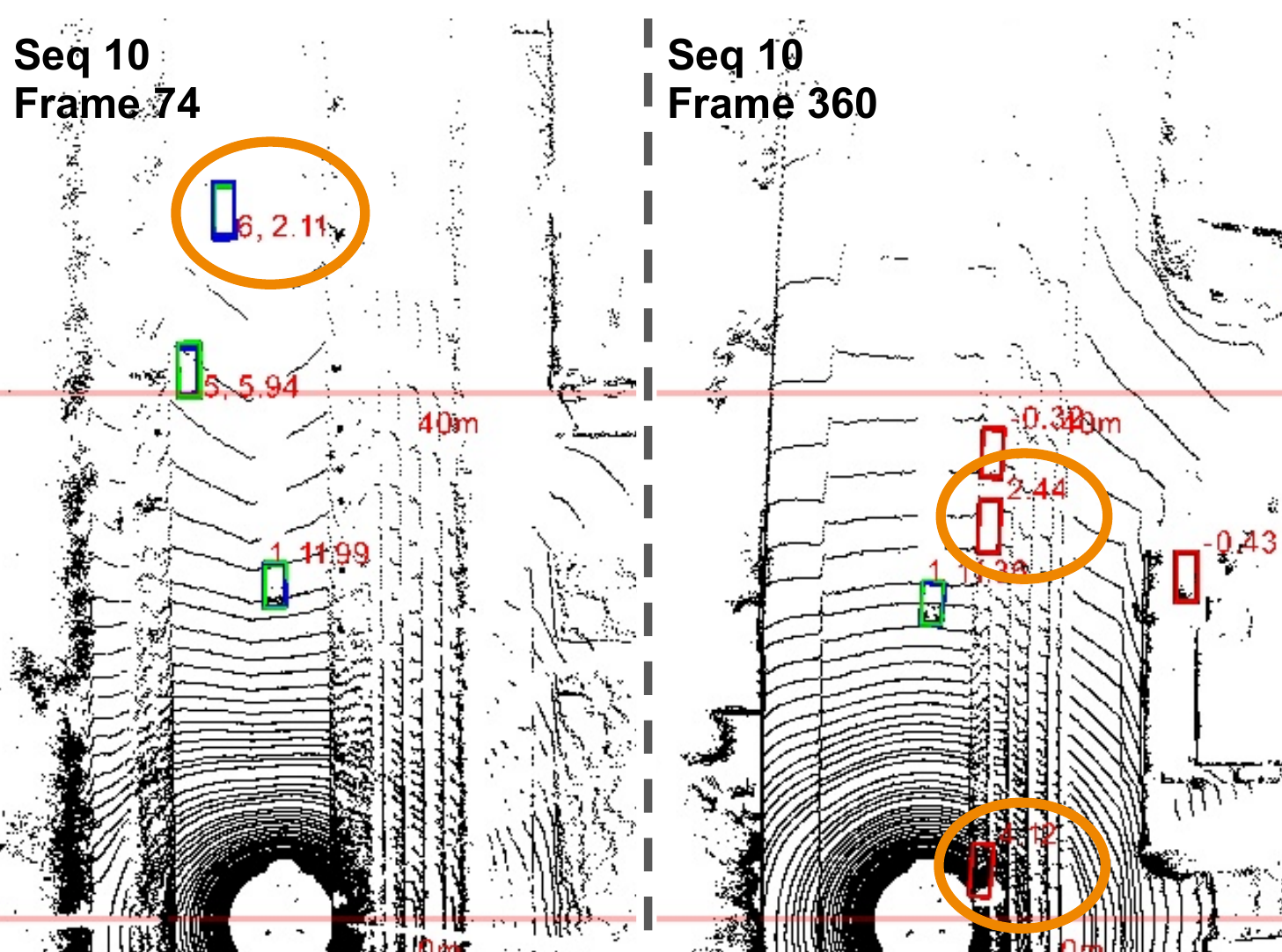}
\end{center}
  \vspace{-0.55cm}
  \caption{Visualization of our 3D MOT system with \textbf{frame-level AutoSelect}. The top/bottom rows show results on images and point clouds respectively. We draw GT objects in \textcolor{blue}{blue}, outputs of our 3D MOT system (\emph{i.e.}, tracked objects) in \textcolor{Green}{green}, and filtered detections in \textcolor{red}{red}. As highlighted in the \textcolor{orange}{orange} ellipses, our AutoSelect can preserve TP detections with lower scores (\emph{e.g.}, score of -0.61 in seq08-frame119) while filter out FP detections with higher scores (\emph{e.g.}, score of 2.70 in seq08-frame225), which demonstrates that our AutoSelect can dynamically search detection threshold suitable in each frame.}
  \label{fig:kitti_qua}
  \vspace{-0.6cm}
\end{figure*}

\section{Experiments}

We evaluate our 3D MOT system on KITTI and nuScenes \cite{Caesar2019}, which provide 2D/3D sensor data to satisfy our need. We do not evaluate on 2D MOT datasets such as MOTChallenges~\cite{Dendorfer2020} as they do not provide 3D sensor data and are not directly applicable to our 3D MOT system. We use standard CLEAR metrics \cite{Bernardin2008} (including MOTA, MOTP, IDS, FRAG, FP, FN) and also the new sAMOTA, AMOTA and AMOTP metrics proposed in \cite{Weng2020_AB3DMOT} for evaluation. We follow the same 3D MOT evaluation protocol as \cite{Weng2020_AB3DMOT} which employs KITTI validation set. This is because KITTI MOT benchmark only supports 2D MOT evaluation on the test set. For KITTI, we report results on the car subset for comparison. For nuScenes, we evaluate on all categories and then compute the mean over all categories. We compare with recent open-source S.O.T.A. 3D MOT systems such as FANTrack \cite{Baser2019}, mmMOT \cite{Zhang2019_robust}, AB3DMOT \cite{Weng2020_AB3DMOT}, GNN3DMOT \cite{Weng2020_GNN3DMOT}. For a fair comparison, we use the same 3D detections for all 3D MOT methods, \emph{i.e.}, \cite{Shi2019} in KITTI and \cite{Zhu2019} in nuScenes. 

\subsection{Comparison with State-of-the-Art Methods}

We summarize 3D MOT evaluation on the KITTI dataset in Table \ref{tab:kitti_quan}. Ours with AutoSelect consistently outperforms others in most of the metrics at different evaluation criteria (\emph{e.g.}, 3D $\text{IoU}_{\text{thres}}$ = 0.25, 0.5, and 0.7). When compared with GNN3DMOT which our system is built on top of, ours with AutoSelect consistently reduce FPs while not obviously increase FNs. Surprisingly, AutoSelect can reduce FNs in some cases (\emph{e.g.}, when $\text{IoU}_{\text{thres}}$ = 0.5). This is because sometimes a FP detection can have a high similarity score with a tracklet, which prevents this tracklet from being matched with its corresponding TP detection. If we filter out such FP detections, then the TP detections can be matched which reduces FNs. As a result, ours establish new S.O.T.A. 3D MOT performance on KITTI and removes the need of efforts for manual threshold search. In addition to evaluation on KITTI, we also evaluate on nuScenes dataset and follow the same evaluation protocol in \cite{Weng2020_GNN3DMOT, Weng2020_AB3DMOT} which uses a minimum distance $\text{Dist}_{\text{thres}}$ of $2$ meters as the 0matching criteria. The results are summarized in Table \ref{tab:nuscenes_quan}. Similar to the trend of results in KITTI, ours with AutoSelect which can reduce FPs largely improves MOTA performance, which further brings improvements in sAMOTA/AMOTA as well.

\begin{table}[t]
\caption{3D MOT Performance on the nuScenes dataset.}
\vspace{-0.3cm}
\centering
\resizebox{\hsize}{!}{
\begin{tabular}{@{}lrrrrr@{}}
\toprule
Method       & Matching criteria & \textbf{sAMOTA} & AMOTA & AMOTP & MOTA \\
\midrule
FANTrack~\cite{Baser2019} & $\text{Dist}_{\text{thres}}$ = 2 & 19.64 & 2.36 & 22.92 & 18.60 \\
mmMOT~\cite{Zhang2019_robust} & $\text{Dist}_{\text{thres}}$ = 2 & 23.93 & 2.11 & 21.28 & 19.82 \\ 
GNN3DMOT~\cite{Weng2020_GNN3DMOT} & $\text{Dist}_{\text{thres}}$ = 2 & 29.84 & 6.21 & 24.02 & 23.53 \\
AB3DMOT~\cite{Weng2020_AB3DMOT} & $\text{Dist}_{\text{thres}}$ = 2 & 39.90 & 8.94 & 29.67 & 31.40 \\
\midrule
\textbf{Ours (Frame-level)} & $\text{Dist}_{\text{thres}}$ = 2 & 46.44 & 11.64 & 35.30 & 41.45 \\
\textbf{Ours (Instance-level)} & $\text{Dist}_{\text{thres}}$ = 2 & \textbf{48.93} & \textbf{12.92} & \textbf{36.99} & \textbf{44.49} \\
\bottomrule
\end{tabular}}
\vspace{-0.43cm}
\label{tab:nuscenes_quan}
\end{table}

\begin{figure}[t]
\begin{center}
\vspace{0.15cm}
\includegraphics[trim=0.1cm 0cm 1.6cm 1.3cm, clip=true, width=0.49\linewidth]{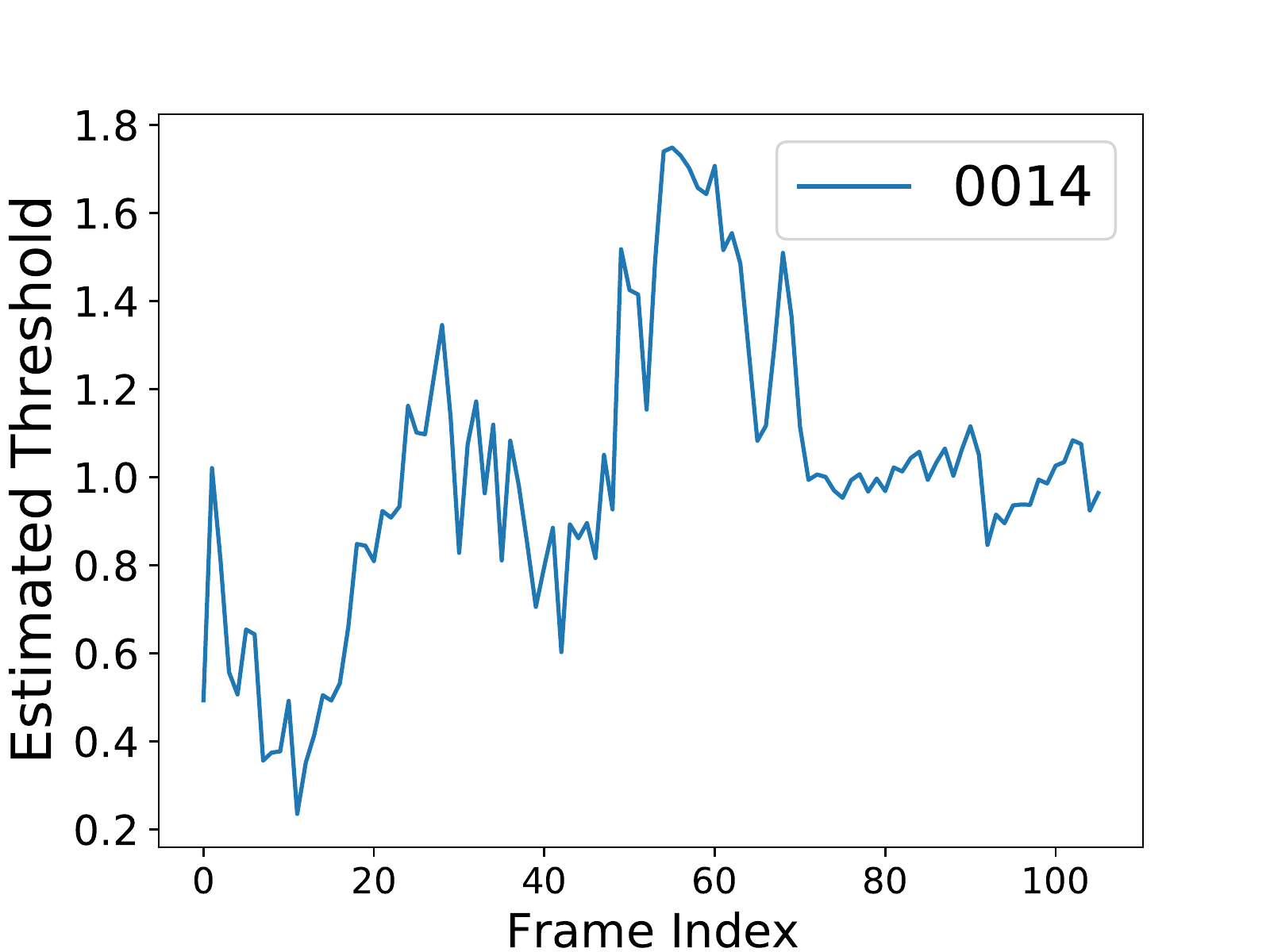}
\includegraphics[trim=0.1cm 0cm 1.6cm 1.3cm, clip=true, width=0.49\linewidth]{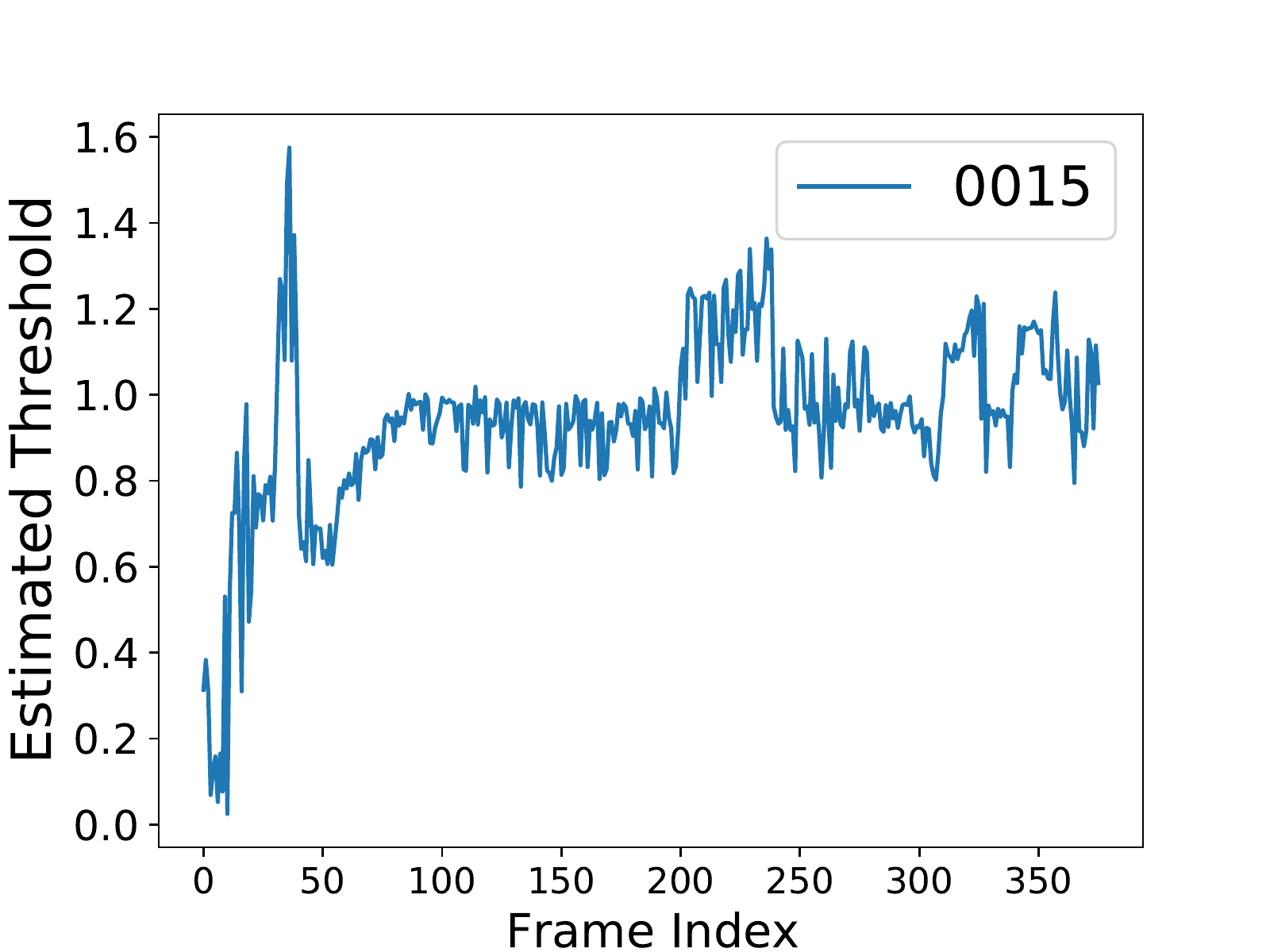}
\end{center}
  \vspace{-0.6cm}
  \caption{We show the estimated detection thresholds by AutoSelect on two KITTI data sequences, which are indeed dynamic over frames.}
  \label{fig:overtime}
  \vspace{-0.6cm}
\end{figure}

\begin{figure}[t]
\begin{center}
\includegraphics[trim=0cm 4.5cm 10.5cm 5cm, clip=true, width=\linewidth]{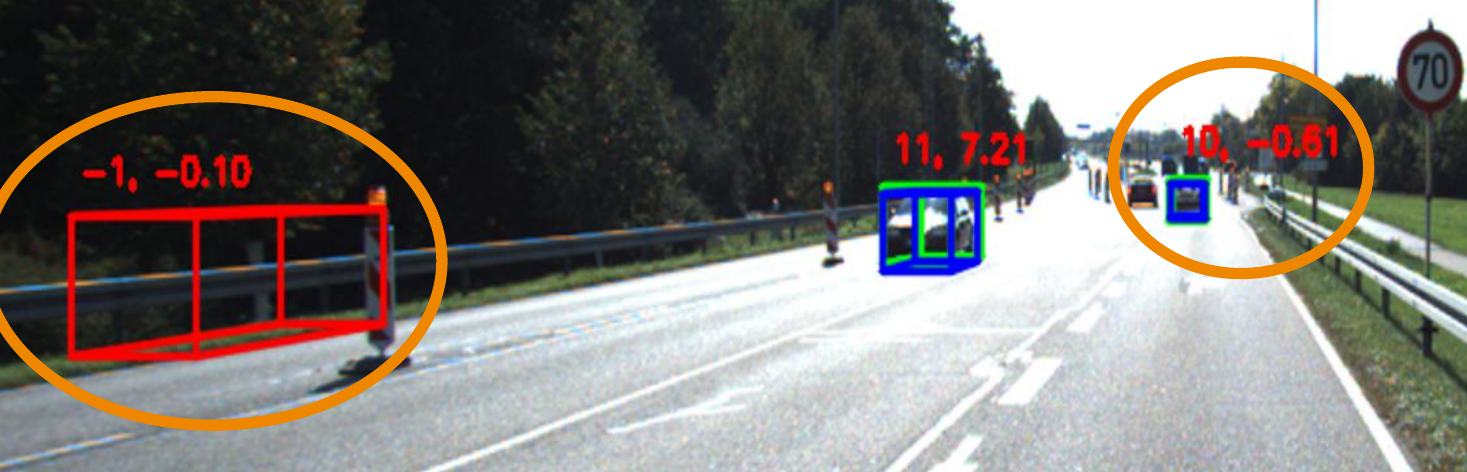}
\includegraphics[trim=4.5cm 5cm 6cm 4.5cm, clip=true, width=\linewidth]{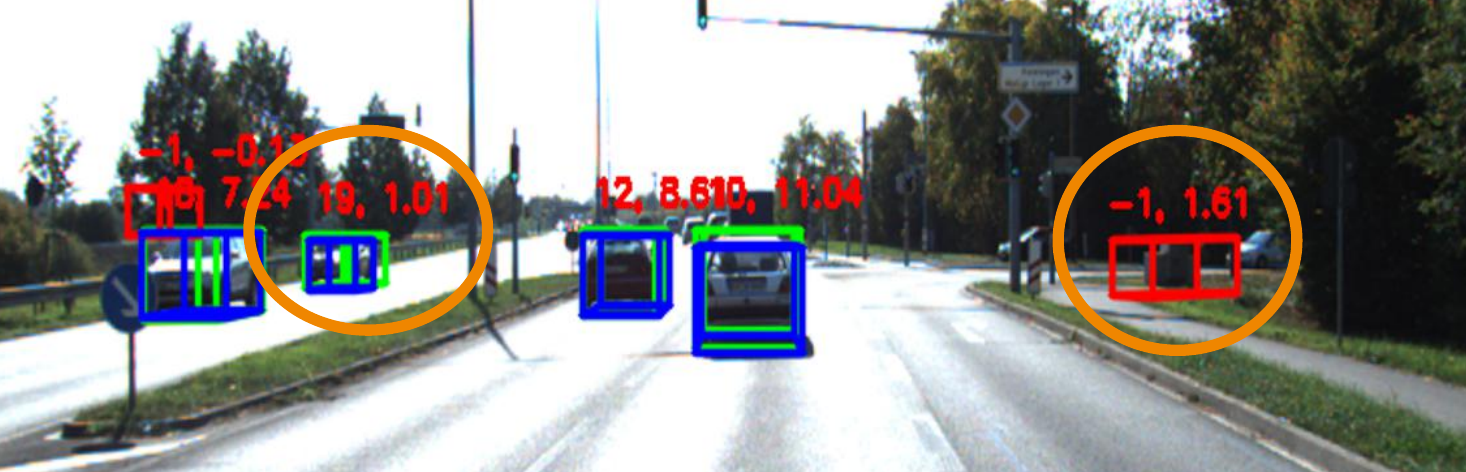}
\end{center}
  \vspace{-0.55cm}
  \caption{Visualization of our 3D MOT system with \textbf{instance-level AutoSelect}. We show that our instance-level AutoSelect can successfully filter out high-score FPs and maintain low-score TPs in the same frame.}
  \label{fig:qua_inst}
  \vspace{-0.6cm}
\end{figure}

\subsection{Analysis}

To demonstrate that frame-level AutoSelect can effectively reduce FP detections while preserve TPs, we visualize results of example frames from three sequences of the KITTI dataset in Fig. \ref{fig:kitti_qua}. As highlighted in orange ellipses, AutoSelect can indeed estimate different thresholds across frames, \emph{i.e.}, estimate a lower threshold to preserve TP detections with low scores in a particular frame while output a higher threshold to filter out FP detections with high scores in another frame. To further confirm how frame-level AutoSelect works, we plot the estimated thresholds over frames in \ref{fig:overtime}, which demonstrates that our estimated threshold can indeed dynamically adapt based on the input data in the current frame and a few recent history frames. 

Although frame-level AutoSelect has demonstrated success, what about the problem shown in Fig. \ref{fig:instance} that frame-level AutoSelect cannot properly separate FP/TP detections if the FP detections' scores are higher than scores of TPs? Is the proposed instance-level AutoSelect able to resolve this issue? To answer the questions, we visualize the results of our 3D MOT system using instance-level AutoSelect in Fig. \ref{fig:qua_inst}. We can see that, although in a same frame, our instance-level AutoSelect can successfully filter out FPs with higher scores and preserve TPs with relatively lower scores, which provides further performance improvements over the frame-level AutoSelect as shown in Table \ref{tab:kitti_quan} and \ref{tab:nuscenes_quan}. 

\section{Conclusions}

We proposed an automatic and dynamic detection selection technique that (1) can be embedded to modern 3D MOT methods for end-to-end training, (2) can remove the need for extensive search of the detection confidence threshold, (3) can dynamically adapt detection selection to a particular frame or to specific object detections and improve 3D MOT performance by more effective FP reduction. As our current method is designed specifically for 3D MOT, an extension to 2D MOT (\emph{e.g.}, embed our AutoSelect with a S.O.T.A. 2D MOT method and show improved performance in video-based MOT benchmarks) will have some practical value.









\bibliographystyle{IEEEtran}
\bibliography{IEEEabrv,main}

\end{document}